\documentclass[11pt,a4paper]{article}
\usepackage[hyperref]{acl2021}
\usepackage{times}
\usepackage{latexsym}
\usepackage{adjustbox}

\usepackage{microtype}

\aclfinalcopy 

\usepackage{times}
\usepackage{latexsym}
\usepackage{multirow}
\usepackage{amsmath,amssymb,amsfonts}
\usepackage{mathtools}
\usepackage{breqn}
\usepackage{bm}      
\usepackage[T1]{fontenc}

\usepackage[utf8]{inputenc}

\usepackage{microtype}
\usepackage{color,soul}
\usepackage{balance}
\usepackage{diagbox}

\title{A Globally Normalized Neural Model for Semantic Parsing}

\author{
  Chenyang Huang$^{1}$, 
  Wei Yang$^2$,
  Yanshuai Cao$^2$,
  Osmar Za\"iane$^{1}$,
  Lili Mou$^{1}$ \\
  $^1$Alberta Machine Intelligence Institute (Amii), University of Alberta\ \ $^2$Borealis AI \\
  \tt{\{chenyangh,zaiane\}@ualberta.ca} \\ \tt{\{wei.yang,yanshuai.cao\}@borealisai.com}  \\
  {\tt doublepower.mou@gmail.com}}

\date{}

\begin{document}
\maketitle
\begin{abstract}
In this paper, we propose a globally normalized model for context-free grammar (CFG)-based semantic parsing. Instead of predicting a probability, our model predicts a real-valued score at each step and does not suffer from the label bias problem. Experiments show that our approach outperforms locally normalized models on small datasets, but it does not yield improvement on a large dataset. 
\end{abstract}

\setlength{\abovedisplayskip}{7pt}
\setlength{\belowdisplayskip}{7pt}
\setlength{\abovedisplayshortskip}{7pt}
\setlength{\belowdisplayshortskip}{7pt}

\section{Introduction}
Semantic parsing has received much interest in the NLP community \cite{ZelleM96,ZettlemoyerC05,jia-liang-2016-data,guo-etal-2020-benchmarking-meaning}. 
The task is to map a natural language utterance to executable code, 
such as $\lambda$-expressions, SQL queries, and Python programs. 

Recent work integrates the context-free grammar (CFG) of the target code into the generation process. 
Instead of generating tokens of the code \cite{dong-lapata-2016-language}, CFG-based semantic parsing predicts the grammar rules in the abstract syntax tree (AST). This guarantees the generated code complies with the CFG, and thus it has been widely adopted \cite{YinN18,guo-etal-2019-towards,bogin-etal-2019-representing,sun2019grammar,SunZXSMZ20}.

Typically, the neural semantic parsing models are trained by maximum likelihood estimation (MLE). The models predict the probability of the next rules in an autoregressive fashion, known as a locally normalized model. 

However, local normalization is often criticized for the label bias problem \cite{laffertycrf,AndorAWSPGPC16,wiseman2016sequence,stanojevic-steedman-2020-max}. In semantic parsing, for example, grammar rules that generate identifiers (e.g., variable names) have much lower probability than other grammar rules. Thus, the model will be biased towards such rules that can avoid predicting identifiers. More generally, the locally normalized model will prefer such early-step predictions that can lead to low entropy in future steps.

In this work, we propose to apply global normalization to neural semantic parsing. 
Our model scores every grammar rule with an unbounded real value, instead of a probability, so that the model does not have to avoid high-entropy predictions and does not suffer from label bias. 
Specifically, we use max-margin loss for training, where the ground truth is treated as the positive sample and beam search results are negative samples. 
In addition, we accelerate training by initializing the globally normalized model with the parameters from a pretrained locally normalized model.

We conduct experiments on three datasets: 
ATIS \cite{dahl-etal-1994-expanding}, CoNaLa \cite{yin2018mining}, and Spider \cite{yu-etal-2018-spider}.
Compared with local normalization, our globally normalized model is able to achieve higher performance on the small ATIS and CoNaLa datasets with the long short-term memory (LSTM) architecture, but does not yield improvement on the massive Spider dataset when using a BERT-based pretrained language model.

\section{Related Work}
Early approaches to semantic parsing mainly rely on predefined templates, and are domain-specific \cite{ZelleM96,ZettlemoyerC05,kwiatkowksi-etal-2010-inducing}. 
Later, researchers apply sequence-to-sequence models to semantic parsing. \newcite{dong-lapata-2016-language} propose to generate tokens along the syntax tree of a program. \newcite{YinN17} generate a program by predicting the grammar rules; our work uses the TranX tool~\cite{YinN18} with this framework.

Globally normalized models, such as the conditional random field \cite[CRF,][]{laffertycrf}, are able to mitigate the label bias problem. However, their training is generally difficult due to the global normalization process. To tackle this challenge, 
\newcite{daumelaso} propose learning as search optimization (LaSO), and \newcite{wiseman2016sequence} extend it to the neural network regime as beam search optimization (BSO).
Specifically, they obtain negative partial samples whenever the ground truth falls out of the beam during the search, and ``restart'' the beam search with the ground truth partial sequence teacher-forced.

Our work is similar to BSO. However, we search for an entire output, and do not train with partial negative samples. This is because our decoder is tree-structured, and different partial trees cannot be implemented in batch efficiently. We instead perform locally normalized pretraining to ease the training of our globally normalized model.

\section{Methodology}

In this section, we first introduce the neural semantic parser TranX, which servers as the locally normalized base model in our work. We then elaborate how to construct its globally normalized version.

\subsection{The TranX Framework}

TranX is a context-free grammar (CFG)-based neural semantic parsing system \cite{YinN18}. 
TranX first encodes a natural language input $X$ with a neural network encoder.

\begin{figure}[!t]
  \includegraphics[width=\linewidth]{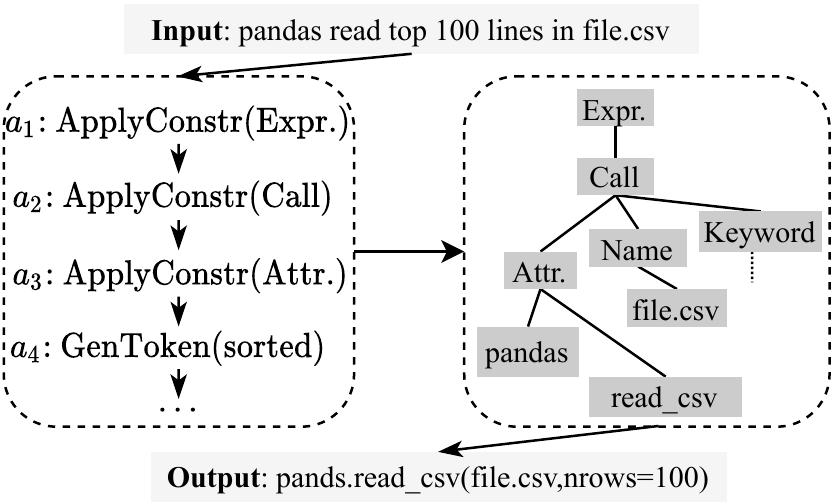}
  \centering
  \caption{An example of generating a Python program with TranX.}
  \label{fig:tranx_ex}
\end{figure}

Then, the model generates a program by predicting the grammar rules (also known as actions) along the abstract syntax tree (AST) of the program. In Figure~\ref{fig:tranx_ex}, for example, the rules generating the desired program include \text{ApplyConstr(Expr.)}, \text{ApplyConstr(Call)}, \text{ApplyConstr(Attr.)}, and \text{GenToken(sorted)}.

In TranX, these actions are predicted in an autoregressive way based on the input $X$ and the partially generated tree, given by
\begin{align}
\resizebox{.95\linewidth}{!}{$
\begin{aligned}
    P_L(a_t |\bm a_{<t}, X ; \bm \theta_L)\!=\! \frac{\exp\{ o(a_t | \bm a_{<t}, X; \bm \theta_L)\}} {\!\!\!\!\sum\limits_{a'_t \in \mathcal{A}_t(\bm a_{<t})}\!\!\!\exp \{ o(a'_t | \bm a_{<t} , X; \bm \theta_L)\}}\label{eq:local_norm}
\end{aligned}
$}
\end{align}
where $\bm \theta_L$ denotes the parameters of the neural network model, and the subscript $L$ emphasizes that the probability is locally normalized. 
$o(\cdot)$ denotes the logit at this step, and $a_t$ is an action (i.e., grammar rule) among all possible actions at this step $\mathcal{A}_t(\cdot)$, which is based on previous predicted rules $\bm a_{<t}$. 

In other words, the prediction probability is normalized at every step, and the training objective is to maximize
\begin{align}
\resizebox{.7\linewidth}{!}{$
\begin{aligned}
    P_L (\bm a_{1:n} | X; \bm \theta_L) = \prod_{t=1}^n P_L(a_t |\bm a_{<t}, X ;\bm \theta_L)
\end{aligned}
$}\label{eq:local}
\end{align}
where $n$ is the total number of steps.

\subsection{Globally Normalized Training}

A locally normalized model may suffer from the label bias problem~\cite{laffertycrf}.  
This is because such a model normalizes the probability to 1 at every step. However, the candidate action set $\mathcal{A}_t(\bm a_{<t})$ may have different sizes,
and the actions from a smaller $\mathcal{A}_t(\bm a_{<t})$ typically have higher probabilities.
Thus, the model would prefer such actions $\bm a_{<t}$ that will yield smaller $\mathcal A_t(\bm a_{<t})$ in future steps.\footnote{Or more generally, the model prefers $\mathcal A_t(\bm a_{<t})$ with a smaller entropy.}  

We propose to adapt TranX to a global normalized  model to alleviate label bias. Instead of predicting a probability $P(a_t|\bm a_{<t}, X)$ as in (\ref{eq:local}), our globally normalized model predicts a positive score at a step as 
\begin{align} 
\resizebox{!}{!}{$
\begin{aligned}
    s(a_t|\bm a_{<t}, X ;\bm\theta_G)= \exp \{o(a_t|\bm a_{<t}, X ;\bm\theta_G) \}
\end{aligned}
$}
\end{align}
where $o(\cdot)$ is the same logit as (\ref{eq:local_norm}), and $\bm \theta_G$ is the parameters.

The probability of the sequence $\bm a_{1:n}$ is normalized only once in a global manner, given by 
\begin{align}
\resizebox{!}{!}{$
\begin{aligned}
    P_G(\bm a_{1:n}, X; \bm \theta_G) = \frac{1}{Z_G} \prod_{t=1}^n s (a_t | \bm a_{<t}, X ; \bm \theta_G)
\end{aligned}
$}\label{eq:global}
\end{align}
where $Z_G = \sum_{\bm a'_{1:n}}\prod_{t=1}^n s (a'_t | \bm a'_{<t} ; \bm \theta_G) $ is the partition function. 

A globally normalized model alleviates the label bias problem, because it does not normalize the probability at every prediction step, as seen from (\ref{eq:global}). Thus, it is not biased by the size of $\mathcal A_t(\bm a_{<t})$. 

The training objective is still to maximize the likelihood, albeit normalized in a global way. 
However, computing the partition function $Z_G$ requires enumerating all combinations of actions $\bm a_{1:n}'$ in the partition function of (\ref{eq:global}), which is generally intractable.

In practice, the maximum likelihood training is approximated by max-margin loss between a positive sample $\bm a_{1:n}$ and a negative sample $\bm a^-_{1:n}$, given by
\begin{align}
\label{eq:global_loss}
\resizebox{0.95\linewidth}{!}{$
\begin{aligned}
 \mathcal{L}(\bm  a^{-}_{1:n}, \bm  a_{1:n} )= \max \{0,  o(\bm  a^{-}_{1:n} | X ) - o(\bm a_{1:n} | X ) + \Delta \}
\end{aligned}
$}
\end{align}
where $o(\bm a_{1:n}|X) = \frac{1}{n} \sum_{t=1}^n o(a_t | \bm a_{<t} ) $ is the average of logits. $\Delta$ is a positive constant. 

The positive sample is simply the ground truth actions, whereas the negative samples are obtained by beam search. In other words, we perform beam search inference during training, and the sequences in the beam (other than the ground truth) serve as the negative samples.

Similar to MLE training for (\ref{eq:global}), the max-margin loss increases the logits of the ground truth sample, while decreasing the logits for others. It is noted that the quality of negative samples will largely affect the max-margin training, as only a few samples are used to approximate $Z_G$.

To address this issue, we initialize the parameters of the globally normalized model $\bm \theta_G$ with $\bm \theta_L$ in a pretrained locally normalized model. Thus, our negative samples are of higher quality, so that the max-margin training is easier and more stable.

\subsection{Handling the Copy Mechanism}
\label{sec:global_copy}
TranX has a copy mechanism \cite{gu-etal-2016-incorporating} as an important component for predicting the terminal nodes of the AST, as the target program largely overlaps with the source utterance, especially for entities (e.g., ``file.csv'' in Figure~\ref{fig:tranx_ex}). In the locally normalized TranX, the copy mechanism marginalizes the probability of generating a token in the vocabulary and copying it from the source: 
\begin{align}
\resizebox{!}{!}{$
\begin{aligned}
    &P_L(a_{t}=\operatorname{GenToken}[v]\,|\,\bm a_{<t}, X ) \label{eq:local_copy}\\ \nonumber
    &= P(\operatorname{gen}\,|\,\bm a_{<t}, X) P(v\,|\,\operatorname{gen}, \bm a_{<t}, X ) \\ \nonumber
    &\ \ + P(\operatorname{copy}\,|\, \bm a_{<t}, X) P(v \,|\, \operatorname{copy}, \bm a_{<t}, X) \nonumber
\end{aligned}
$}
\end{align}
where $\operatorname{GenToken}[\cdot]$ denotes generating a terminal token $v$. $P(\operatorname{copy}| \cdot)$ is the predicted probability of copying the token $v$ from the source utterance, and $P(\operatorname{gen} | \cdot) = 1 - P(\operatorname{copy} | \cdot )$ is the probability of generating $v$ from the vocabulary.

However, the copy mechanism cannot be directly combined with global normalization, because we use unbounded, real-valued logits instead of probabilities. This would not make much sense when both logits are negative, whereas their product is positive. 

Therefore, we propose a variant of copy mechanisms in the globally normalized setting.
Specifically, we keep the probabilities $P(\operatorname{copy}|\cdot)$ and $P(\operatorname{gen}|\cdot)$, and use them to 
weight the logits of generating and copying a token $v$, given by 
\begin{align}
\resizebox{.7\linewidth}{!}{$
\begin{aligned}
    &o(a_{t}=\operatorname{GenToken}[v]\,|\,\bm a_{<t}, X ) \\ \nonumber
    &= P(\operatorname{gen}\,|\,\bm a_{<t}, X) o(v\,|\,\operatorname{gen}, \bm a_{<t}, X ) \\ \nonumber
    &\ \ + P(\operatorname{copy}\,|\, \bm a_{<t}, X) o(v \,|\, \operatorname{copy}, \bm a_{<t}, X) \nonumber
\end{aligned}
$}
\end{align}

Here, $o(a_{t}=\operatorname{GenToken}[v]\,|\,\cdot )$ is a linear interpolation of two logits, and thus fits the max-margin loss (\ref{eq:global_loss}) naturally.

\section{Experiments}

\textbf{Datasets.} We conduct experiments on three benchmark parsing datasets: \href{https://worksheets.codalab.org/bundles/0xbb06bd9208c8436f9277fe202c6ad651}{ATIS} \cite{zettlemoyer-collins-2007-online},  \href{https://conala-corpus.github.io}{CoNaLa} \cite{yin2018mining}, and \href{https://yale-lily.github.io/spider}{Spider} \cite{yu-etal-2018-spider}, which contain 4473, 2379, and 8695 training samples, respectively.

It should be pointed out that much work adopts data anonymization techniques to replace entities with placeholders \cite{dong-lapata-2016-language,YinN17,yin-neubig-2019-reranking,SunZXSMZ20}. This unfortunately causes a large number of duplicate samples between training and test. This is recently realized in~\newcite{guo-etal-2020-benchmarking-meaning}, and thus, in our work, we only compare the models using the original, correct ATIS dataset. 

\textbf{Settings.} Our globally normalized semantic parser is developed upon the open-sourced TranX\footnote{\url{https://github.com/pcyin/tranX}}.
We adopt the CFG grammars provided by TranX to convert lambda calculus and Python programs into ASTs and sequence of grammar rules (actions). 
For ATIS and CoNaLa datasets, we use long LSTM models as both the encoder and the decoder. Their dimensions are set to 256. 
For the Spider dataset, we use a pretrained BERT model\footnote{Specifically, we use the RoBERTa-base model \cite{liu2019roberta} as we find it performs better than the original BERT-base model \cite{devlin-etal-2019-bert}.} \cite{devlin-etal-2019-bert} and the relation-aware Transformer \cite{wang-etal-2020-rat} as the encoder and an LSTM as the decoder. The architecture generally follows the work by \newcite{xuoptimizing}.

The beam size is set to 20 to search for negative samples, and is set to 5 for inference. 
The margin $\Delta$ in (\ref{eq:global_loss}) is set to 0.1.  We use the Adam optimizer \cite{KingmaB14} with a learning rate of 5e-4 for training. 

For both ATIS and CoNaLa datasets, 
we report the best results on the development sets and the corresponding results on the test set. For the Spider dataset, we only report the results on the development set as the ground truth of the test set is not publicly available.

\begin{table}[t]
\centering
\resizebox{0.9\linewidth}{!}{
\begin{tabular}{lll}  \hline \hline
        & Dev     & Test    \\ \hline 
\newcite{jia-liang-2016-data}\\
\ \ No copy     &  N/A     &   69.90\%       \\
\ \ Copy & N/A     & 76.30\% \\
\ \ Copy + data recombination     &  N/A     &  \textbf{83.30}\% \\
\hline
\newcite{guo-etal-2020-benchmarking-meaning}\\
\ \  No copy & N/A   &  68.70\%   \\ 
\ \  Copy         &  N/A     &  75.70\%  \\ \hline
Ours  &        &       \\
\ \ No copy                   & 80.00\% & {71.49}\% \\
\ \ Copy  & 79.15\% & 75.63\% \\
\ \ Copy + global              & 81.61\% & 76.32\% \\ 
\   \ Copy + global + emb             & \textbf{84.53}\% & \textbf{78.16}\% \\ \hline \hline
\end{tabular}}
\caption{Exact match accuracy on the ATIS dataset.}
\label{tab:atis}
\end{table}

\section{Results}

\textbf{ATIS dataset.}
Following \newcite{YinN17,SunZXSMZ20}, we report the exact match accuracy for ATIS.
We first replicate locally normalized models with and without the copy mechanism and achieve similar results to \newcite{jia-liang-2016-data} and \newcite{guo-etal-2020-benchmarking-meaning}, shown in Table~\ref{tab:atis}. This verifies that we have a fair implementation and are ready for the study of global normalization. 

We observe that the copy mechanism largely affects the accuracy on the test set, although it has little effect on the development set. This is because the training and validation distributions closely resemble each other, whereas the test distribution differs largely. Therefore, the copy mechanism is important for handling unseen entities in the test set, and our proposed copy variant in Section~\ref{sec:global_copy} is also essential to globally normalized models.

We then train our model with the max-margin loss. Our globally normalized model consistently improves the accuracy on both development and test sets, compared with its locally normalized counterpart. This shows the effectiveness of our approach.

In addition, we notice that a large number of entities in ATIS have a form like ``ap:denvor'' (Denver Airport). We thus use the combination of character-level ELMo embeddings \cite{peters-etal-2018-deep} and word-level GloVe embeddings \cite{pennington-etal-2014-glove}. This further improves the accuracy, which outperforms the previous methods by $\sim\!\!1.9\%$ in the setting without data augmentation.

\textbf{CoNaLa dataset.} For CoNaLa,
BLEU is treated as the main metric in previous work \cite{yin-neubig-2019-reranking}, because accuracy is generally very low (<3\%) on this dataset. 
From Table~\ref{tab:conala}, we observe that our globally normalized model improves the BLEU scores on both the development and test sets compared with the locally normalized baseline. Such improvement is consistent with that on ATIS.

We further compare our model with \newcite{yin-neubig-2019-reranking}, which reranks beam search results by heuristics. Our method is outperformed by the reranking approach. Note that reranking can be considered as alleviating label bias with postprocessing, as the locally normalized model fails to assign the correct sequence with the highest joint probability. 
However, the reranking method requires training several reranking scorers, combined with an ad hoc feature (namely, length). By contrast, our global normalization does not rely on ad hoc human engineering.

\begin{table}[!t]
\centering
\resizebox{0.85\linewidth}{!}{
\begin{tabular}{lllll} \hline \hline
                           & Dev & Test \\  \hline 
\newcite{YinN18}        &       N/A       & 24.35\%  \\
~~~~ + Reranking &      N/A       & 30.11\%  \\ \hline 
Ours (local)              & 33.46\% & 25.84\%  \\
~~~~ + Reranking              & \textbf{35.82}\% & \textbf{28.39}\%  \\
~~~~ + Global                & 34.75\% & 27.08\% \\  \hline \hline
\end{tabular}}
\caption{BLEU score on the CoNaLa dataset.}
\label{tab:conala}
\end{table}
\begin{table}[!t]
\centering
\resizebox{0.75\linewidth}{!}{
\begin{tabular}{lllll} \hline \hline
                           & Dev Acc.   \\ \hline 
\newcite{rubin2020smbop}  &    73.4\%  \\
\newcite{yu2021grappa}      &   74.7\% \\ \hline 
Ours  (local)               & 73.79\% \\
\ \ + Global   &  73.69\% \\   \hline   \hline 
\end{tabular}}
\caption{Exact match accuracy on the Spider dataset. Test performance requires submissions to the official website. We report validation performance instead.}
\label{tab:spider}
\end{table}

\textbf{Spider dataset.}
Table~\ref{tab:spider} lists the results on the Spider dataset. 
Here, our locally normalized model uses BERT as the encoder, 
and its performance is on par with that from the recent state-of-the-art approaches \cite{rubin2020smbop,yu2021grappa}. However, our global normalization does not improve the performance. 
It is noted that BERT is a more powerful model than LSTM, and Spider has a much larger training set than CoNaLa and ATIS.
We conjuncture that BERT learns the step-by-step local prediction probability very well, which in turn  yields a satisfying joint probability and largely mitigates label bias by itself.
Therefore, the globally normalized model does not exhibit its superiority on the Spider dataset.

\section{Conclusion}

In this work, we propose to apply global normalization for neural semantic parsing.
Our approach predicts the score of different grammar rules at an autoregressive step, and thus it does not suffer from the label bias problem. We observe that our proposed method is able to improve performance on small datasets with LSTM-based encoders. However, global normalization becomes less effective on the large dataset with a BERT architecture.

\section*{Acknowledgments}

We acknowledge the support of the Natural Sciences and Engineering Research Council of Canada (NSERC) under grant Nos.~RGPIN-2020-04465 and RGPIN-2020-04440.
Chenyang Huang is supported by the Borealis AI Graduate Fellowship Program.
Lili Mou and Osmar Za\"{\i}ane are supported by the Amii Fellow Program and the Canada CIFAR AI Chair Program. This research is also supported in part by Compute Canada (\url{www.computecanada.ca}).

\bibliographystyle{acl_natbib}
\bibliography{custom}

\end{document}